\documentclass[onecolumn, 12pt, final]{IEEEtran}
\IEEEoverridecommandlockouts
\usepackage{cite}
\usepackage{amsmath,amssymb,amsfonts, amsthm}
\usepackage{algorithmic}
\usepackage{algorithm}
\usepackage{graphicx}
\usepackage{svg}
\usepackage{textcomp}
\usepackage{xcolor}
\usepackage{mathtools}
\usepackage{booktabs}
\usepackage{multirow}
\usepackage{longtable}
\usepackage{listings}
\usepackage{soul} %
\usepackage{subfigure}
\usepackage{caption}
\usepackage{tikz}
\usepackage[colorlinks]{hyperref}
\usepackage[acronym,nonumberlist]{glossaries}
\usepackage[capitalize]{cleveref}
\usepackage{subcaption}
\usepackage[section]{placeins}

\makeglossaries
\usepackage{framed,enumitem}

\hypersetup{
    colorlinks  = true,
    linkcolor   = blue,     %
    anchorcolor = black,
    citecolor   = blue,   %
    filecolor   = cyan,
    menucolor   = red,
    runcolor    = cyan,
    urlcolor    = blue    %
}
\newcommand{\etal}{et al.~}

\theoremstyle{definition}

\newacronym{nserc}{NSERC}{Natural Sciences and Engineering Research Council of Canada}
\newacronym{oce}{OCE}{Ontario Center of Excellence}
\newacronym{ppe}{PPE}{Personal Protective Equipment}
\newacronym{osh}{OSH}{Occupational Safety and Health}
\newacronym{dl}{DL}{Deep Learning}
\newacronym{od}{OD}{Object Detection}
\newacronym{cv}{CV}{Computer Vision}
\newacronym{cnn}{CNN}{Convolutional Neural Network}
\newacronym{yolo}{YOLO}{You Only Look Once}
\newacronym{covid}{COVID-19}{Coronavirus Disease}
\newacronym{gpu}{GPU}{Graphical Processing Unit}
\newacronym{rcnn}{R-CNN}{Region Proposal Convolutional Neural Network}
\newacronym{fpn}{FPN}{Feature Pyramid Network}
\newacronym{cspnet}{CSPNet}{Cross-Stage Partial Connections}
\newacronym{panet}{PANet}{Path Aggregation Network}
\newacronym{sam}{SAM}{Spatial Attention Module}
\newacronym{ciou}{CIoU}{Complete Intersection Over Union}
\newacronym{silu}{SiLU}{Sigmoid-weighted Linear Unit}
\newacronym{elan}{ELAN}{Efficient Long-Range Attention Network}
\newacronym{gelan}{GELAN}{Generalized Efficient Long-Range Attention Network}
\newacronym{repconvn}{RepConvN}{Reparameterized Convolutional}
\newacronym{dfl}{DFL}{Distribution Focal Loss}
\newacronym{lfyolo}{LF-YOLO}{Lighter and Faster YOLO}
\newacronym{rmf}{RMF}{Reinforced Multiscale Feature}
\newacronym{attyolo}{ATT-YOLO}{Attention YOLO}
\newacronym{senet}{SENet}{Squeeze and Excitation Network}
\newacronym{yoloimf}{YOLO-IMF}{YOLO for Industrial Manufacturing Field}
\newacronym{mscoco}{MS-COCO}{Microsoft Common Objects in Context}
\newacronym{rtx}{RTX}{Ray Tracing Texel eXtreme}
\newacronym{p}{P}{Precision}
\newacronym{r}{R}{Recall}
\newacronym{tp}{TP}{True Positive}
\newacronym{fp}{FP}{False Positive}
\newacronym{tn}{TN}{True Negative}
\newacronym{fn}{FN}{False Negative}
\newacronym{map}{mAP}{Mean Average Precision}
\newacronym{iou}{IoU}{Intersection over Union}
\newacronym{flops}{FLOPS}{Floating-Point Operations Per Second}
\newacronym{id}{ID}{Identifier}
\newacronym{deepsort}{DeepSort}{Deep Simple Object Tracking}
\newacronym{kf}{KF}{Kalman Filtering}
\newacronym{mes}{MES}{Manufacturing Execution System}
\newacronym{lmm}{LMM}{Large Multimodal Model}
\newacronym{ssim}{SSIM}{Structural Similarity Index Measure}
\newacronym{tsne}{t-SNE}{T-distributed Stochastic Neighbor Embedding}
\newacronym{fps}{fps}{Frames Per Second}
\newacronym{ml}{ML}{Machine Learning}
\newacronym{iot}{IoT}{Internet of Thing}
\newacronym{roi}{RoI}{Region of Interest}
\newacronym{ssd}{SSD}{Single Stage Detector}
\newacronym{nms}{NMS}{Non-Max-Suppression}
\newacronym{pgi}{PGI}{Programmable Gradient Information}
\newacronym{ai}{AI}{Artificial Intelligence}

\begin{document}
\title{SH17: A Dataset for Human Safety and Personal Protective Equipment Detection in Manufacturing Industry
\thanks{
\IEEEauthorrefmark{1} Corresponding author.\\
This study is supported by IFIVEO CANADA INC., Mitacs through IT16094, \acrfull{nserc} through ALLRP 560406-20, \acrfull{oce} through OCI\# 34166, and the University of Windsor, Canada.
}}
\author{
    \IEEEauthorblockN{Hafiz Mughees Ahmad\IEEEauthorrefmark{2}\IEEEauthorrefmark{1} and Afshin Rahimi\IEEEauthorrefmark{4}} \\
    \IEEEauthorblockA{Mechanical, Automotive and Materials Engineering Department, \\
    University of Windsor, Windsor, ON, Canada\\
    \IEEEauthorrefmark{2}ahmad54@uwindsor.ca, \IEEEauthorrefmark{4}arahimi@uwindsor.ca}
}
\maketitle

\begin{abstract}
Workplace accidents continue to pose significant risks for human safety, particularly in industries such as construction and manufacturing, and the necessity for effective \acrfull{ppe} compliance has become increasingly paramount. Our research focuses on the development of non-invasive techniques based on the \acrfull{od} and \acrfull{cnn} to detect and verify the proper use of various types of \acrshort{ppe} such as helmets, safety glasses, masks, and protective clothing.
This study proposes the SH17 Dataset, consisting of 8,099 annotated images containing 75,994 instances of 17 classes collected from diverse industrial environments, to train and validate the \acrshort{od} models. We have trained state-of-the-art \acrshort{od} models for benchmarking, and initial results demonstrate promising accuracy levels with \acrfull{yolo}v9-e model variant exceeding 70.9\% in \acrshort{ppe} detection.
The performance of the model validation on cross-domain datasets suggests that integrating these technologies can significantly improve safety management systems, providing a scalable and efficient solution for industries striving to meet human safety regulations and protect their workforce. The dataset is available at \url{https://github.com/ahmadmughees/sh17dataset}.
\end{abstract}

\begin{IEEEkeywords}
SH17, Object Detection, Convolutional Neural Network, YOLO, Personal Protective Equipment, Worker, Human Safety, Dataset
\end{IEEEkeywords}

\section{Introduction} \label{sec:introduction}
\acrfull{osh} is a multidisciplinary field that ensures safety in work environments and avoids risks to a worker’s health. This includes preventing work-related injuries, accidents, and illnesses by managing hazards and promoting healthful practices. \acrshort{ppe} is crucial in industrial manufacturing for safeguarding workers against various occupational hazards. The importance of \acrshort{ppe} lies in its ability to minimize the risks of injuries and illnesses from exposure to dangerous conditions such as toxic chemicals, extreme temperatures, and mechanical hazards. Research emphasizes that \acrshort{ppe} is an essential last line of defense in protecting workers when other safety measures are insufficient \cite{suttonChapterPersonalProtective2015}.

Based on the guidelines from the Occupational Safety and Health Administration (OSHA) \cite{occupationalsafetyandhealthadministrationoshaPersonalProtectiveEquipment2004}, and Vukicevic \etal \cite{isailovicComplianceHeadmountedIndustrial2022}, this study divides \acrshort{ppe} into five categories according to the body parts it protects; 1) Head 2) Upper Body 3) Hands 4) Feet and 5) Whole body. Common examples of these \acrshort{ppe}s include helmets, safety glasses, face shields, earmuffs, safety vests, gloves, and safety shoes, each designed to address specific threats. Studies have shown that proper selection, usage, and maintenance of \acrshort{ppe} are vital for ensuring its effectiveness and the safety of the employees.

Manufacturing industries are trying their best to ensure the safety of their workforce and use of \acrshort{ppe} in their facilities to reduce the injuries causing millions of dollars in damages. Earlier systems use sensor-based approaches for automatic \acrshort{ppe} compliance but with the recent popularity of \acrfull{ml} and \acrfull{cv} systems, researchers have been proposing non-invasive solutions for \acrshort{ppe} compliance for the manufacturing facilities \cite{nathDeepLearningSite2020, yuComplexRealWorldSafety2023}. Vukicevic \etal \cite{vukicevicGenericComplianceIndustrial2022} proposed 2 stage approach for the detection of \acrshort{ppe} using person and keypoint detection in the first step and then detecting the \acrshort{ppe} type in the second stage using \acrshort{od} model. Yu \etal \cite{yuComplexRealWorldSafety2023} proposed to use the \acrfull{yolo}v5 \acrshort{od} model for detecting \acrshort{ppe} in chemical plants.

In this study, we also propose to use \acrshort{od}-based methods for the \acrshort{ppe} compliance. To make this study more accessible and help other researchers pursue advancement in this field, we offer an open-source dataset consisting of 8,099 images and 75,994 instances. We compare this dataset with existing benchmark datasets and relevant studies and provide open-source weights for \acrshort{ppe} detection to make it easy for others to recreate the results.

The main contributions of this work are,
\begin{itemize}
    \item A collection of a high-quality, large-scale dataset from the internet. The dataset contains images from all over the world, removing the location and gender bias and making it inclusive. The dataset is open-sourced for commercial and research purposes.
    \item Extensive training of the state-of-the-art \acrshort{od} models with publicly available weights for community use.
    \item Evaluation of trained model with different dataset to test the efficacy across out-of-distribution data.
\end{itemize}

The remainder of this paper is structured as follows: a literature review and background are provided in \cref{sec:literature_review}. The collected dataset is introduced in \cref{sec:sh17}. The insights from the experiments and results are provided in \cref{sec:experiment_and_discussion}. Finally, \cref{sec:conclusion} offers concluding remarks and future directions.

\section{Literature Review} \label{sec:literature_review}
Human safety inspectors are still the most common and easiest way to ensure \acrshort{ppe} compliance at any workplace. However, it is costly and laborious work with a high error margin. Earlier, the wide usage of \acrfull{iot} devices in industrial manufacturing introduced sensors for \acrshort{ppe} compliance in the process. However, that is also a very costly and sensitive process \cite{naticchiaMonitoringSystemRealtime2013, kelmMobilePassiveRadio2013}. Recently, \acrfull{cv} methods have emerged as a non-invasive solution, providing cheaper and better alternatives to the sensors \cite{isailovicComplianceHeadmountedIndustrial2022, xiangFastRobustSafety2024, ahmadCapacityConstraintAnalysis2024} and well accepted in construction as well as manufacturing. These methods use the annotated data to train the \acrshort{ml} models to detect the required objects.
In the following section, we have discussed the existing datasets for \acrshort{ppe} detection and \acrshort{od} methods used in the literature.

\subsection{Existing Datasets} \label{subsec:existing_datasets}

Most existing datasets for \acrshort{ppe} detection are tailored towards hardhat (often called helmets) detection in construction engineering.

\subsubsection{Hardhat Wearing Detection (GDUT-HWD) Dataset}
The GDUT-HWD dataset \cite{wuAutomaticDetectionHardhats2019a} contains hardhat images collected from internet sources in 5 different colors labeled individually. It contains 3,174 images and 18,893 object instances within multiple sizes. Due to crowd-sourcing, its publicly available version\footnote{\url{https://github.com/wujixiu/helmet-detection}} contains a lot of advertisement images with non-relevant backgrounds.

\subsubsection{Safety Helmet Wearing (SHW) Dataset}
The SHW dataset \cite{njvisionpowerNjvisionpowerSafetyHelmetWearingDataset2024} is a comprehensive public dataset\footnote{\url{https://github.com/njvisionpower/Safety-Helmet-Wearing-Dataset}} used for detecting both safety helmets and heads. It comprises 7,581 images featuring 9,044 instances of humans wearing safety helmets and 111,514 instances of humans without safety helmets. The examples were sourced from search engines and a portion from the SCUT-HEAD dataset \cite{pengDetectingHeadsUsing2018} and manually annotated afterward.

\subsubsection{Color Helmet and Vest (CHV) Dataset}
The CHV dataset \cite{wangFastPersonalProtective2021} is publicly available\footnote{\url{https://github.com/ZijianWang-ZW/PPE_detection}} and consists of 1,330 high-quality images labeled into 6 categories of person, vests, and colored helmets in 4 colors. The authors selected images from GDUT-HWD \cite{wuAutomaticDetectionHardhats2019a} and SHW \cite{njvisionpowerNjvisionpowerSafetyHelmetWearingDataset2024} datasets using the strict criteria of related construction background, gestures of workers, and object angles and distances from the camera.

\subsubsection{SHEL5K Dataset}
The SHEL5K dataset \cite{otgonboldSHEL5KExtendedDataset2022} is an improved version of Safety Helmet Detection (SHD) dataset\footnote{\url{https://www.kaggle.com/datasets/andrewmvd/hard-hat-detection}} \cite{SafetyHelmetDetection} consisting of 5,000 images with 75,570 instances in 6 classes, including head, helmet, face, and person with and without a helmet. The dataset is publicly available\footnote{\url{https://data.mendeley.com/datasets/9rcv8mm682/4}}.

\subsubsection{Pictor-PPE Dataset}
The Pictor-PPE dataset \cite{nathDeepLearningSite2020} is one of the earliest datasets collected for the \acrshort{ppe} detection. It contains 3 classes, Hat, Vest, and Worker, and contains 774 crowd-sourced and 698 web-mined images. Its publicly available version\footnote{\url{https://github.com/ciber-lab/pictor-ppe}} only contains 784 images available for scientific purposes from construction sites with a total of 2,496 worker instances.

\subsubsection{TCRSF Dataset}
The TCRSF dataset \cite{yuComplexRealWorldSafety2023} is a closed-source dataset\footnote{Authors announced in \cite{yuComplexRealWorldSafety2023} that dataset will be publicly released at \url{https://github.com/sofffty/TCRSF}. However, it has not been released as of 2024-04-19.} and extracted from the video feeds collected from different viewpoints in the chemical plants. It provides more realistic scenarios from complex backgrounds. It contains 50,558 labeled instances from 12,373 images in 7 categories, including helmet, safety clothing, head, etc\footnote{Information mentioned in their paper \cite{yuComplexRealWorldSafety2023}}.

All the above datasets, summarized in \cref{tab:existing_datsets}, are collected to address the safety needs of the construction industry in particular, but human safety in the general manufacturing environment is not fully considered, and in this study, we have explored this holistic aspect by collecting data from diverse environments.

\begin{table}[htbp!]
\centering
\normalsize
\caption{Existing Datasets of helmet and \acrshort{ppe} detection.}
\label{tab:existing_datsets}
\begin{tabular}{lcrrcc}
\toprule
Dataset                            & Classes & Images & Instances & Available                                                          & Paper \\ \midrule
Pictor-PPE                         & 3       & 784    & -     & \href{https://github.com/ciber-lab/pictor-ppe}{\checkmark}                        & \cite{nathDeepLearningSite2020} \\ 
SHW        & 1       & 7,581   & 120,558   & \href{https://github.com/njvisionpower/Safety-Helmet-Wearing-Dataset}{\checkmark} & \cite{njvisionpowerNjvisionpowerSafetyHelmetWearingDataset2024} \\ 
CHV        & 6       & 1,330   & -     & \href{https://github.com/ZijianWang-ZW/PPE_detection}{\checkmark}                 & \cite{wangFastPersonalProtective2021} \\ 
TCRSF                              & 7       & 12,373  & 50,558     & \href{https://github.com/sofffty/TCRSF}{$\times$}                               & \cite{yuComplexRealWorldSafety2023} \\ 
GDUT-HWD & 5       & 3,174   & 18,893     & \href{https://github.com/wujixiu/helmet-detection}{\checkmark}                    & \cite{wuAutomaticDetectionHardhats2019a} \\ 
SHD      & 3       & 5,000      & -         & \href{https://www.kaggle.com/datasets/andrewmvd/hard-hat-detection}{\checkmark}   & - \\ 
SHEL5K                             & 5       & 5,000   & 75,570     & \href{https://data.mendeley.com/datasets/9rcv8mm682/4}{\checkmark}                & \cite{otgonboldSHEL5KExtendedDataset2022} \\ \midrule
SH17 (proposed)               & 17       & 8,099   & 75,994     & \href{https://github.com/ahmadmughees/SH17dataset}{\checkmark} & - \\ \midrule
\end{tabular}

\end{table}

\subsection{YOLO Models} \label{subsec:yolos}
\acrfull{yolo} is a series of popular real-time object detection models that has been through several iterations from \acrshort{yolo}v1 to \acrshort{yolo}v8 \cite{redmonYouOnlyLook2016, redmonYOLO9000BetterFaster2017, redmonYolov3IncrementalImprovement2018,bochkovskiyYOLOv4OptimalSpeed2020,jocherUltralyticsYolov5V32020, liYOLOv6SingleStageObject2022, wangYOLOv7TrainableBagoffreebies2023}. The latest version, \acrshort{yolo}v8, builds upon its predecessors by incorporating state-of-the-art techniques and innovations to improve accuracy, speed, and adaptability.

\acrshort{yolo}v8 \cite{jocherUltralyticsYolov5V32020} follows a one-stage \acrshort{od} approach, where the input image undergoes a single pass through the network for bounding box prediction and classification. The architecture consists of three main components: a backbone network, a neck, and a prediction head following redmon \etal\cite{redmonYolov3IncrementalImprovement2018} and bochkovskiy \etal \cite{bochkovskiyYOLOv4OptimalSpeed2020}. While the previous variants used the same head for objectness, classification, and regression tasks. Jocher \etal proposed the decoupled head along with \acrfull{ciou} loss \cite{zhengDistanceIoULossFaster2020} and \acrfull{dfl} \cite{liGeneralizedFocalLoss2020}, especially for the calculation of bounding box loss. 

The major drawback of \acrshort{yolo}v8 was reliance on DarkNet-53\cite{redmonYolov3IncrementalImprovement2018}. It limited the ability to capture fine-grained features, especially for small and occluded objects. Wang \etal proposed \acrshort{yolo}v9 \cite{wangYOLOv9LearningWhat2024} to address this by introducing \acrfull{pgi}, a novel concept that ensures the retention of critical information throughout the detection process. \acrshort{pgi} integrates a reversible branch that works alongside the main network, preserving essential features and improving training outcomes without additional computational costs. Additionally, they proposed a \acrfull{gelan} block that optimizes the balance between parameter count, complexity, accuracy, and inference speed, enabling users to select the best computational blocks for various devices.

Recently, proposed \acrshort{yolo}v10 \cite{wangYOLOv10RealTimeEndtoEnd2024} further builds upon predecessors and improves the inference speed by eliminating the non-max-suppression post-processing step which only picks the most probable bounding box out of 1,000s of boxes produced by \acrshort{od} model. The authors proposed a consistent dual assignment strategy where a one-to-one assignment head is introduced in addition to a one-to-many head \cite{chenDATEDualAssignment2022} with an identical structure. Both heads are optimized together during training utilizing rich features, while only a one-to-one head is used in the inference for better inference speed. They also reduced the number of trainable parameters by carefully analyzing each trainable block and removing non-contributing blocks to the overall efficiency. They achieved comparable performance as of \acrshort{yolo}v9-c variant by reducing the 46\% trainable parameters. 

The next section discusses the use of these \acrshort{od} methods in the \acrshort{ppe} detection for human safety.

\subsection{\acrfull{od} for \acrshort{ppe} Detection}
Most previous studies that focus on using the \acrshort{od} for \acrshort{ppe} detection have considered only the applications in construction engineering aiming to verify the use of hard hats and safety vests.
Isailovic \etal \cite{isailovicComplianceHeadmountedIndustrial2022} and Vukicevic \etal \cite{vukicevicGenericComplianceIndustrial2022} proposed the \acrshort{ppe} compliance using a two-stage approach by using a keypoint detector to detect regions and then passing these regions to the binary classification or \acrshort{od} model for further \acrshort{ppe} detection. They used the images from Pictor-PPE \cite{nathDeepLearningSite2020} and hardhat dataset \cite{xieHardhat2019} to achieve this objective.

Wu \etal \cite{wuAutomaticDetectionHardhats2019} used the \acrfull{ssd} \cite{liuSSDSingleShot2016} architecture to identify hardhats of different colors on construction sites. They collected the novel GDUT-HWD dataset to train the model.
Delhi \etal \cite{delhiDetectionPersonalProtective2020} and Tran \etal \cite{tranFullyAutomatedVisionbased2019} similarly employed \acrshort{yolo}v3 to detect hardhats and safety jackets in the construction environment and further used it to check for \acrshort{ppe} compliance on various body parts \cite{tranFullyAutomatedVisionbased2019} labeling safe and not-safe to each worker. They gathered the data from web sources.
Otgonbold \etal \cite{otgonboldSHEL5KExtendedDataset2022} compared the performance of multiple \acrshort{od} models for detecting 6 different classes from person, helmet, head, and face on the novel SHEL5k dataset \cite{otgonboldSHEL5KExtendedDataset2022}.
Nath \etal \cite{nathDeepLearningSite2020} used different combinations of \acrshort{od} approaches to address worker safety and \acrshort{ppe} compliance. The authors initially detected the hard, vest, and worker and passed these to \acrfull{ml} classifier, while in another approach, they trained the model to predict the state of the \acrshort{ppe}. They passed the cropped \acrshort{roi} for a person to another \acrshort{cnn} classifier for \acrshort{ppe} classification. The Pictor-PPE dataset \cite{nathDeepLearningSite2020} is publicly available.
Chen and Demachi \cite{chenVisionBasedApproachEnsuring2020} introduced a method using OpenPose \cite{caoRealtimeMultiperson2d2017} for body landmark detection and \acrshort{yolo}v3 \cite{redmonYolov3IncrementalImprovement2018} \acrshort{od} model for \acrshort{ppe} detection. They analyzed the geometric relationships between the key points and detected \acrshort{ppe} to assess compliance. They used the GDUT-HWD dataset for all their experiments.
Zhafran \etal \cite{zhafranComputerVisionSystem2019} explored the Fast \acrshort{rcnn} \cite{girshickFastRcnn2015} architecture for checking masks, gloves, hardhats, and vests, noting a decrease in accuracy with greater distance and changes in lighting conditions using the manually collected laboratory data.
Additionally, several studies have focused on detecting protective masks due to the \acrfull{covid} pandemic.
Loey \etal \cite{loeyFightingCOVID19Novel2021} used \acrshort{yolo}v2 while \cite{nagrathSSDMNV2RealTime2021} combined \acrshort{ssd} \cite{liuSSDSingleShot2016} with MobileNetV2 \cite{sandlerMobileNetV2InvertedResiduals2018} for medical mask detection during \acrshort{covid}.

Ferdous and Ahsan \cite{ferdousPPEDetectorYOLObased2022} used \acrshort{yolo}X \cite{geYOLOXExceedingYOLO2021} \acrshort{od} model on the CHV dataset. Kim \etal \cite{kimApplicationYOLOV52023} scrapped the internet to collect a novel dataset of 4,844 images and manually annotated them into 3 classes, heavy, \acrshort{ppe}, and worker to train \acrshort{yolo}v5 \cite{jocherUltralyticsYolov5V32020} and \acrshort{yolo}v8 \cite{jocherYOLOUltralytics2023}. Some other relevant works using \acrshort{od} for \acrshort{ppe} detection includes Lung and Wang \cite{lungApplyingDeepLearning2023}, Xiang \etal \cite{xiangFastRobustSafety2024}, Di \etal \cite{diMARAYOLOEfficientMethod2024}, Han \etal \cite{hanUtilizingSyntheticImages2024}, and Azizi \etal \cite{aziziComparisonMachineLearning2024}.

Some traditional approaches include combining the Histogram of Oriented Gradients with the Circle Hough Transform algorithms \cite{rubaiyatAutomaticDetectionHelmet2016}. Li \etal \cite{liAutomaticSafetyHelmet2017} implemented a radiomics-based method for helmet detection, while Mneymneh \etal \cite{mneymnehVisionBasedFrameworkIntelligent2019} introduced a motion detection-based system that subsequently identifies workers and hardhats. Balakrishnan \etal \cite{balakreshnanPPEComplianceDetection2020} designed a software system that includes an \acrshort{iot} module and Microsoft Azure's Custom Vision \acrshort{ai} and Intelligent \acrshort{ai} Services to detect safety glasses in lab settings. Amazon has also introduced the proprietary Amazon Rekognition \acrshort{ppe} detection system \cite{amazonwebservicesDetectingPersonalProtective}.

It is evident that most studies have been motivated by the needs of the construction industry, which remains one of the least digitized sectors and records a high number of fatal injuries. While these studies typically focus on specific types of \acrshort{ppe} such as helmets, vests, and masks, the needs of other industries remain largely unexplored. As per the authors' knowledge, there is no comprehensive study on \acrshort{ppe} compliance for manufacturing study. Instead, separate studies have targeted specific \acrshort{ppe} types relevant to particular industries. These studies typically used a single architecture on custom or small-scale data, making direct comparisons challenging due to the different datasets used for training.

This is the first study to encompass and directly benchmark recent deep learning object detection architectures, providing an objective comparison using a newly developed dataset. This object detection method aims to ensure more efficient compliance with multiple \acrshort{ppe}s on body regions. This study proposes a modular framework for \acrshort{ppe} compliance that can be applied to various types of \acrshort{ppe} and body parts.

\section{SH17 Dataset} \label{sec:sh17}
Current datasets for \acrshort{ppe} detection often focus on specific scenarios, such as detecting helmets, and may not reflect the variety of situations found in real-world industrial settings. In this study, we propose \textbf{\underline{S}}afe \textbf{\underline{H}}uman dataset consisting of \textbf{\underline{17}} different objects referred to as \textbf{\underline{SH17 dataset}}. We scrapped images from the Pexels\footnote{\url{https://www.pexels.com/}} website, showcasing a range of human activities across diverse industrial operations. Samples of the dataset are shown in \cref{fig:sh17_samples}. The following sections discuss data collection, annotation, and further details about the curation process for this dataset.

\begin{figure*}[!htbp]
\input{images/collage/collage_code}
\caption{Samples from the proposed \textbf{SH17 Dataset.} Best viewed online. }
\label{fig:sh17_samples}
\end{figure*}

\subsection{Collection Process}
Many open-source datasets are sourced from platforms such as Flickr\footnote{\url{https://www.flickr.com/}} or gathered through web crawling with search engines such as Google or Bing. However, such data can be noisy and require substantial cleaning effort, with images often subject to different licensing conflicts, resulting in studies not publicly sharing their data. To streamline this process, we gathered images from Pexels, which offers clear usage rights for all its images. To extract relevant images, we used multiple queries such as \textit{manufacturing worker, industrial worker, human worker, labor, etc}. The tags associated with Pexels images proved reasonably accurate. After removing duplicate samples, we obtained around 11,000 samples, of which around 26\% were empty, containing no objects from our target classes. These empty images were excluded during the labeling process, resulting in a final dataset of 8,099 images. The dataset exhibits significant diversity, representing manufacturing environments globally, thus minimizing potential regional or racial biases.

\subsection{Annotation Process}
The annotation process involved four human annotators, three initially completing annotations. The team lead then verified and corrected any mistakes in the annotations. Finally, a graduate student performed a final verification and addressed any remaining mislabeling. All annotators utilized DarkLabel\footnote{\url{https://github.com/darkpgmr/DarkLabel}} while graduate student used the LabelImg\footnote{\url{https://github.com/HumanSignal/labelImg}} for data annotation. The selection of the tool is a personal preference and has no impact on the quality of annotation.

\subsection{Categories}
We categorized the objects into 17 classes, each representing a body part and the required \acrshort{ppe} items for safety. We chose these classes with the aim that downstream applications could ignore irrelevant classes while still covering a wide range of scenarios in industrial manufacturing. The complete list of classes and their definitions can be found in \cref{tab:classes_and_description}. Additional tags with each class to match some of the classes in earlier datasets were also mentioned with each annotation and are released as the extended version of the dataset. The tags are also mentioned in the \cref{tab:classes_and_description}. These can be utilized during training in downstream tasks by advanced users. 
We also extracted the metadata of each image sample. Its details are explained in \cref{sec:meta-data}. 

\begin{table*}[htbp!]
\centering
\normalsize
\caption{List of all the annotated classes and their counts. Tag \textbf{\underline{off}} means the item is present in the scene but not worn by the person, while \textbf{\underline{on}} means the item is present and worn by the person.}
\label{tab:classes_and_description}
\normalsize%
\setlength\tabcolsep{3pt}
\begin{tabular}{lp{2cm}p{2cm}rp{10cm}}
\toprule
ID & Name & Additional Tags & Instances & Description \\ \midrule
1 & Person & male, female, children & 13802 & Uses visible features for classification. \\ \midrule
2 & Head & - & 11985 & Includes any view of the head: front, back,  top or else. \\ \midrule
3 & Face & - & 8950 & Only classified as visible when the nose is visible. \\ \midrule
4 & Glasses & on, off, safety, vision & 1945 & Detection of safety glasses. \\ \midrule
5 & Face-mask-medical & on, off & 669 & Detection of medical face masks \\ \midrule
6 & Face-guard & on, off & 134 & Detection of whether faceguards \\ \midrule
7 & Ear &   & 7730 & Focused on the ears for safety equipment detection. \\ \midrule
8 & Earmuffs & on, off & 318 & Detection of earmuffs. Over-Ear-Headphones are labelled as earmufs.\\ \midrule
9 & Hands &   & 15850 & Focus on the hands for safety equipment detection. \\ \midrule
10 & Gloves & on, off & 2790 & Detection of gloves. \\ \midrule
11 & Foot &   & 796 & Visible when there are no shoes, and each foot is annotated regardless of with or without socks. \\ \midrule
12 & Shoes & on, off, safety, other & 4560 & Shoes detection. Safety: Includes safety shoes and thick joggers. Others: slippers, sneakers, or other types of footwear. \\ \midrule
13 & Safety-vest & on, off & 530 & Detection of safety vests. \\ \midrule
14 & Tools & on, off & 4647 & Detection of tools being held; on means in hand. If the tool is present in the scene but not in hand, it's off. Pencils and laptops are not considered tools. \\ \midrule
15 & Helmet & on, off, white, red, black, yellow, blue & 927 & tags contain the color of a helmet as well.  \\ \midrule
16 & Medical-suit    & on, off & 155 & Detection of medical suits  \\ \midrule
17 & Safety-suit     & on, off & 530 & Detection of safety suits \\
\bottomrule
\end{tabular}

\end{table*}

\subsection{Data Details} \label{subsec:data_details}
\begin{figure*}
    \centering
    \includegraphics[width=1\linewidth]{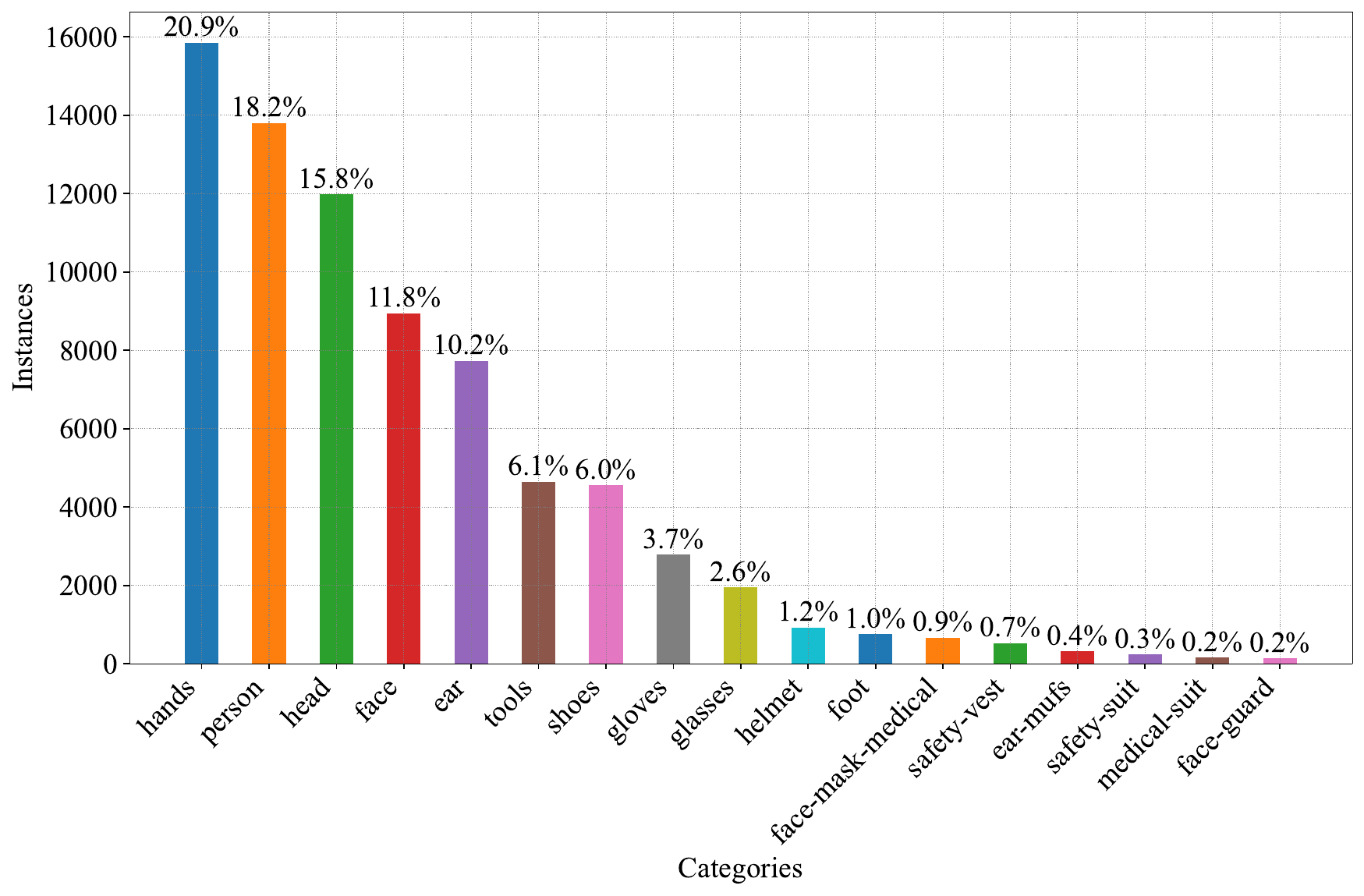}
    \caption{Distribution of all class instances. The percentage of each class is shown at the top of each bar.}
    \label{fig:instance_distribution}
\end{figure*}

We provide the original images extracted from Pexels in their native resolution with a maximum image size of 8,192$\times$5,462 and a minimum of 1,920$\times$1,002. Data consists of both landscape and portrait-style images. Each image contains an average of 9.38 instances. We have labeled ears and earmuffs, which are very small objects compared to a person. Hence, the dataset contains objects of all sizes. There are 39,764 annotations containing less than 1\% of the area, while 59,025 annotations contain less than 5\% area of the image. Examples of all object sizes are included in \cref{fig:sh17_samples}. The dataset contains the maximum instances of hands 15,850, which is 20.9\% of the instances. The helmet class contains 927, approximately 1.2\% of the data, while there are 134 instances of faceguards, the lowest class instance with 0.2\% data only. \Cref{tab:classes_and_description} contains the complete list of instances of each class. \Cref{fig:instance_distribution} shows the distribution of all classes, their instances, and their percentage in the dataset, demonstrating an imbalance in class distribution.

\subsection{Evaluation Metrics} \label{subsec:evaluation_metrics}
Object detection models are evaluated using several metrics to assess their performance accurately. Common metrics used by \acrfull{mscoco} dataset \cite{linMicrosoftCOCOCommon2015} include \acrfull{p}, \acrfull{r}, and \acrfull{map}.

\acrfull{p} measures the proportion of correctly identified objects among all objects detected by the model and is calculated as
\begin{equation*}
\text{\acrlong{p}} = \frac{\text{\acrlong{tp}}}{\text{\acrlong{tp}} + \text{\acrlong{fp}}}
\end{equation*}

\acrfull{r}, on the other hand, quantifies the ability of the model to detect all relevant objects in the dataset and is computed as
\begin{equation*}
\text{\acrlong{r}} = \frac{\text{\acrlong{tp}}}{\text{\acrlong{tp}} + \text{\acrlong{fp}}}
\end{equation*}

Additionally, \acrfull{map} evaluates the detection accuracy across all classes. It's computed based on \acrshort{p} and \acrshort{r} for each class and then averaged to give an overall score. The \acrfull{iou} metric is used to gauge the accuracy of object localization. It calculates the overlap between the ground truth bounding boxes ($b_{g}$) and the model's predicted bounding boxes ($b_{pred}$) as follows:
\begin{equation*}
\text{IoU} = \frac{\text{Area}(b_{pred} \cap b_{g})}{\text{Area}(b_{pred} \cup b_{g})}
\end{equation*}
where $b_{g}$ represents the ground truth bounding box, and $b_{pred}$ denotes the bounding box predicted by the \acrshort{od} model. The \acrshort{iou} threshold acts as a filter to remove \acrshort{fp}) bounding boxes with an \acrshort{iou} score below a certain threshold. This threshold determines the required accuracy for an object to be classified as detected or missed (e.g., \acrshort{iou} $\geq$ threshold). Different models may use various threshold values, such as 0.25, 0.5, or 0.75, during evaluation. 
\section{Experimentation and Discussion} \label{sec:experiment_and_discussion}
We have trained multiple \acrshort{od} models with different size variants of \acrshort{yolo}v8, \acrshort{yolo}v9 and \acrshort{yolo}v10 for benchmarking purposes. We split the dataset into 80\% for training and 20\% for testing. We have used default hyper-parameters suggested by the respective authors in their open-source codebase\footnote{\url{https://github.com/ultralytics/ultralytics}}. We used transfer learning to adapt a pre-trained model initially trained on the \acrshort{mscoco} dataset and trained it on the SH17 dataset for 200 epochs. We followed \cite{jocherYOLOUltralytics2023} for training hyperparameters. We trained all models on the fixed image size of 640 instead of the original high-quality images due to the memory limitation by 2x NVIDIA RTX GPUs with a batch size of 128 for the nano models to 32 for bigger scale models. We also used a mosaic of 4 images along with horizontal flipping for data augmentation and \acrfull{nms} for post-processing the outputs of \acrshort{yolo}v8 and \acrshort{yolo}v9 model variants. We have consolidated the results of the model training in  \cref{tab:consolidated_results} evaluated on the separate test set. As mentioned, \acrshort{yolo}v9-e performs best among all variants with 70.9\% \acrshort{map}@50 and 48.7\% \acrshort{map}50-95. It has 58.1M parameters as compared to the next best \acrshort{yolo}v8-x, which has 68.2M parameters, effectively 15\% fewer parameters. Furthermore, \acrshort{yolo}v9-c has comparable performance to \acrshort{yolo}v8-l with 32\% fewer trainable parameters. \acrshort{yolo}v10-x has comparable performance with \acrshort{yolo}v9-c with 15\% less trainable parameters, which significantly reduces the training and inferences time. 

\begin{figure}[htbp!]
    \centering
    \includegraphics[width=0.95\linewidth]{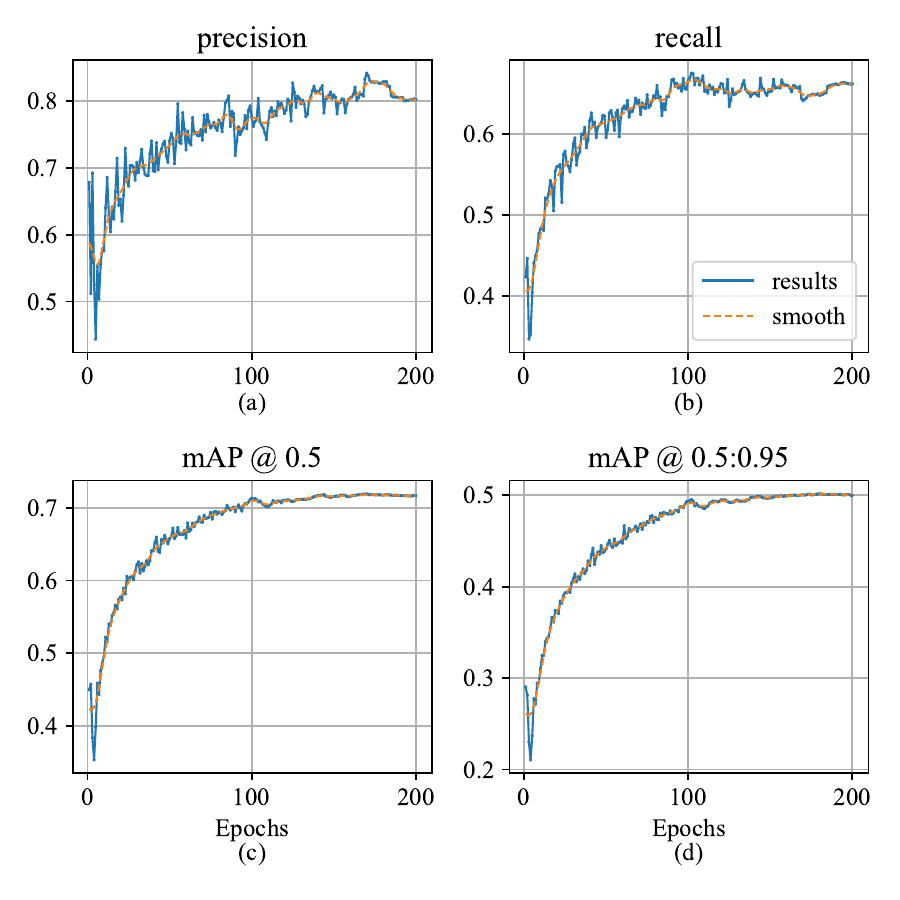}
    \caption{Training metrics of \acrshort{yolo}v9-e model that performs best among all.}
    \label{fig:yolov9-e_training_metrics}
\end{figure}

We present the training metrics of \acrshort{yolo}v9-e model, our best-performing model on SH17, in \cref{fig:yolov9-e_training_metrics}. The training of the model plateaus at around 170 epochs as the \acrshort{map}50 does not improve after that. \cref{tab:yolov9e_results} lists the class-wise accuracy of \acrshort{yolo}v9-e model on the randomly sampled test set. Hands class, having the most samples, performs well with 89.8\% \acrshort{map}. Tools and Foot class samples vary in types, safety shoes, slippers, joggers, and runners, while tools can be anything a person is working with, resulting in these classes showing low \acrshort{map}. 

\begin{table}[htbp!]
\centering
\normalsize
\caption{Comparison of the models trained on SH17. \textbf{Bold} represents the best (a larger value is better).}
\normalsize%
\setlength\tabcolsep{3pt}
\label{tab:consolidated_results}
\begin{tabular}{lrcccccc}
\toprule

   &  & & & &  & \multicolumn{2}{c}{\acrshort{map}} \\
\cmidrule{7-8}
Model   & Params &Images & Instances & P             &  R            & 50             & 50-95    \\ \midrule
Yolo-8-n & 3.2    & 1620  &  15358    & 67.5          & 53.6          & 58.0           &  36.6    \\
Yolo-8-s & 11.2   & 1620  &  15358    & 81.5          & 55.7          & 63.7           &  41.7    \\
Yolo-8-m & 25.9   & 1620  &  15358    & 77.1          & 60.5          & 66.6           &  45.7    \\
Yolo-8-l & 43.7   & 1620  &  15358    & 76.7          & 62.9          & 68.0           &  47.0    \\
Yolo-8-x & 68.2   & 1620  &  15358    & 77.1          & 63.1          & 69.3           &  47.2    \\ \midrule 
Yolo-9-t & 2.0    & 1620  &  15358    & 75.0          & 52.6          & 58.5           &  37.5    \\
Yolo-9-s & 7.2    & 1620  &  15358    & 73.6          & 60.2          & 65.3           &  42.9    \\
Yolo-9-m & 20.1   & 1620  &  15358    & 77.4          & 62.0          & 68.6           &  46.5    \\
Yolo-9-c & 25.5   & 1620  &  15358    & 79.6          & 60.8          & 67.7           &  46.5    \\
Yolo-9-e & 58.1   & 1620  &  15358    & \textbf{81.0} & \textbf{65.0} & \textbf{70.9}  &  \textbf{48.7}    \\ \midrule 
Yolo-10-n & 2.3   & 1620  &  15358    & 66.8          & 53.2          & 57.2           &  35.9    \\
Yolo-10-s & 7.2   & 1620  &  15358    & 75.8          & 57.0          & 62.7           &  40.9    \\
Yolo-10-m & 15.4   & 1620  &  15358    & 71.4          & 61.4          & 65.7           &  43.8    \\
Yolo-10-b & 19.1   & 1620  &  15358    & 77.7          & 59.1          & 65.8           &  45.1    \\
Yolo-10-l & 24.4   & 1620  &  15358    & 76.0          & 61.8          & 67.4           &  46.0    \\
Yolo-10-x & 29.5   & 1620  &  15358    & 76.8          & 62.8          & 67.8           &  46.7    \\
\bottomrule
\end{tabular}

\end{table}

\begin{table}[htbp!]
\centering
\normalsize
\caption{Class-wise accuracy of \acrshort{yolo}v9-e model. }
\label{tab:yolov9e_results}
\normalsize%
\setlength\tabcolsep{3pt}
\begin{tabular}{lrrcccccc}
\toprule

   &  & & & &  \multicolumn{2}{c}{\acrshort{map}} \\
   \cline{6-7}
        Class  & Images & Instances &  \acrshort{p} &  \acrshort{r} &   50 &  50-95  \\ \midrule
         all &      1620  &    15358 &     81.0 &     65.0  &    70.9  &    48.7 \\
       hands &      1284  &     3212 &     91.4 &     83.9  &    89.8  &    64.8 \\
      person &      1515  &     2734 &     90.9 &     89.2  &    92.1  &    77.9 \\
        head &      1314  &     2427 &     94.8 &     89.1  &    93.5  &    74.3 \\
        face &      1155  &     1855 &     96.0 &     88.1  &    93.8  &    73.8 \\
         ear &       987  &     1612 &     91.2 &     75.5  &    84.3  &    55.0 \\
       shoes &       320  &      956 &     79.2 &     62.8  &    70.8  &    43.2 \\
        tool &       455  &      923 &     67.4 &     39.1  &    43.2  &    27.6 \\
      gloves &       254  &      529 &     81.6 &     58.9  &    66.5  &    43.5 \\
     glasses &       323  &      398 &     87.4 &     72.6  &    76.4  &    46.9 \\
      helmet &        93  &      154 &     81.3 &     67.8  &    77.0  &    57.6 \\
   face-mask &        75  &      151 &     88.8 &     73.2  &    75.5  &    49.2 \\
        foot &        64  &      149 &     51.7 &     22.1  &    29.3  &    14.0 \\
 safety-vest &        45  &       97 &     66.4 &     55.0  &    57.7  &    38.1 \\
    ear-mufs &        38  &       49 &     79.6 &     46.9  &    57.1  &    40.5 \\
 safety-suit &        28  &       45 &     65.7 &     53.3  &    58.5  &    38.3 \\
medical-suit &        30  &       43 &     86.0 &     65.1  &    68.5  &    40.6 \\
  face-guard &        23  &       24 &     76.8 &     62.5  &    71.7  &    42.8 \\
\bottomrule
\end{tabular}

\end{table}

\begin{figure*}[htbp]
\input{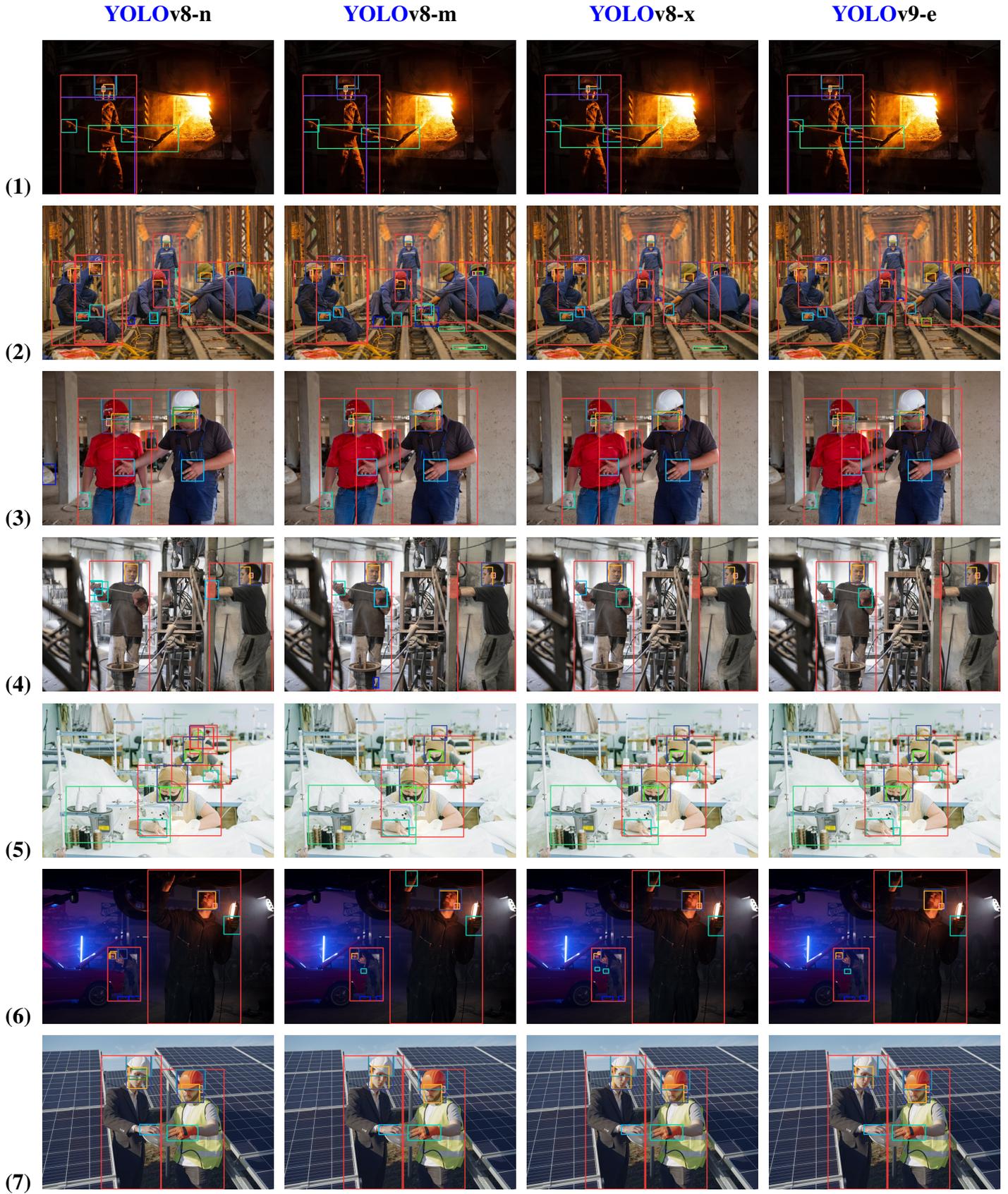}
\caption{Visual representation of the predictions made by different models; \acrshort{yolo}v8-n, \acrshort{yolo}v8-m, \acrshort{yolo}v8-x, \acrshort{yolo}v9-e. Best viewed online.}
\label{fig:model_predictions}
\end{figure*}

These benchmarks further underscore the results reported in the \cite{jocherYOLOUltralytics2023,wangYOLOv9LearningWhat2024,wangYOLOv10RealTimeEndtoEnd2024 } where \acrshort{yolo}v9 models perform comparably with reduced parameters due to the usage of \acrshort{pgi} and \acrshort{gelan}. The \acrshort{yolo}v10 struggles on the small objects and SH17 consists of 52\% objects covering less than 1\% of the area(as explained in \cref{subsec:data_details}).  \Cref{fig:model_predictions} presents some visual differences between the models; \acrshort{yolo}v8-n, \acrshort{yolo}v8-m, \acrshort{yolo}v8-x, \acrshort{yolo}v9-e. For sample (2) in \cref{fig:model_predictions}, \acrshort{yolo}v8-m and \acrshort{yolo}v8-x both made a \acrshort{fp} prediction of the tool on the ground. For sample (3), \acrshort{yolo}v8-n model \acrshort{fp} of face class. In row 5, all models failed to detect the tool in the worker's hand. The original labeled samples are mentioned in the \cref{fig:sh17_samples}.

\subsection{Generalization capability}
\begin{figure}[htbp!]
    \centering
    \includegraphics[width=1\linewidth]{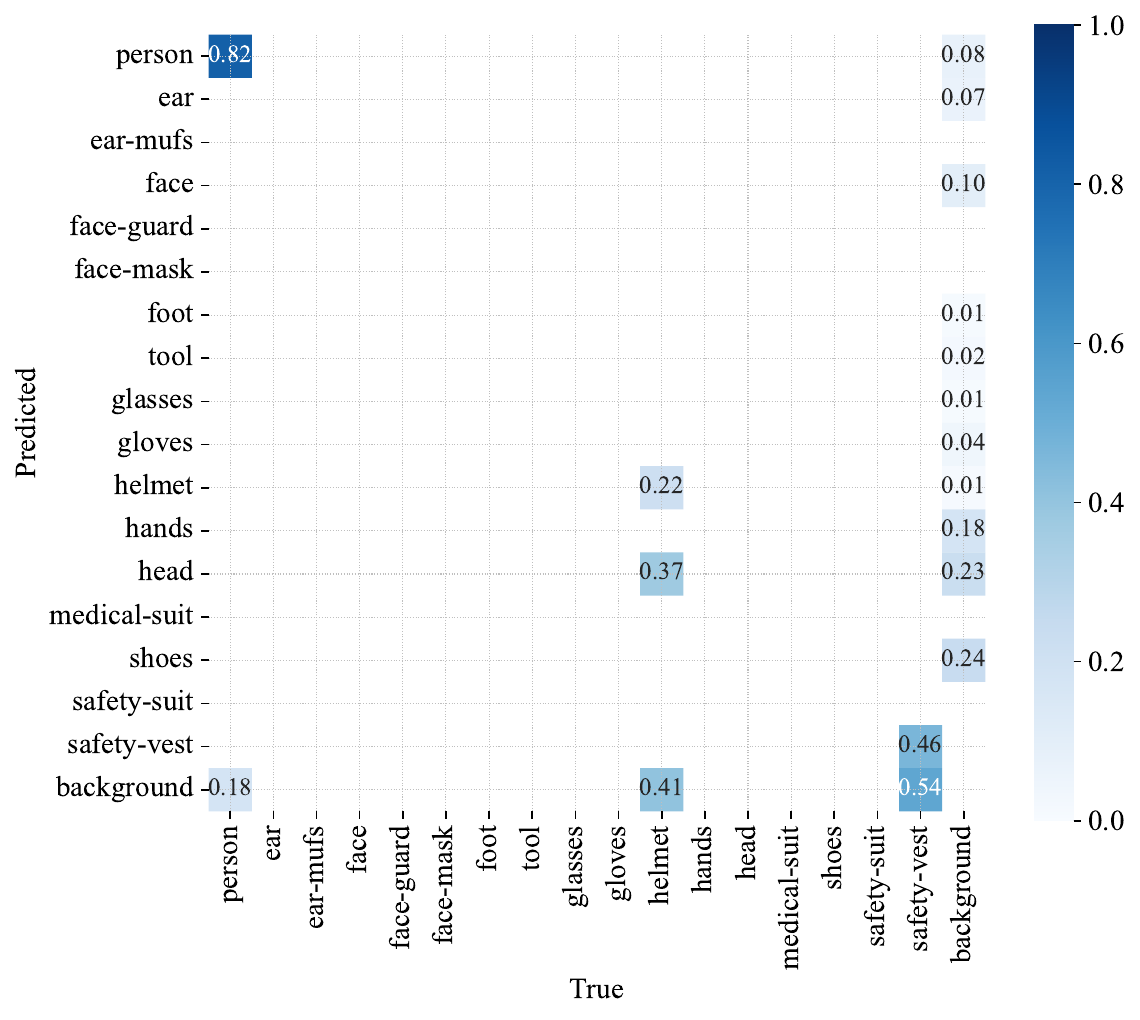}
    \caption{Confusion Matrix of all the instances of Pictor-PEE dataset. }
    \label{fig:conf_mat_pictor_ppe}
\end{figure}

We have verified the model's efficacy with the SH17 dataset on the cross-domain datasets. We selected the Pictor-PPE dataset \cite{nathDeepLearningSite2020} to validate the \acrshort{yolo}v9-e model. The dataset has only three classes: worker, hat, and vest, which we mapped to the Person, Helmet, and Safety-vest classes of SH17. We used all the publicly available images in the dataset for validation, and \acrshort{yolo}v9-e achieved 58.8\% \acrshort{map}, which shows that trained models can be used directly for \acrshort{ppe} compliance in the manufacturing environment. SH17 has the least samples of safety-vest class; however, it still achieves 35.5\% \acrshort{map}50. The complete results are summarised in \cref{tab:yolo_v9e_inference_on_pictor_ppe}. \Cref{fig:conf_mat_pictor_ppe} presents the confusion matrix of the complete Pictor-PPE as just the evaluation set to \acrshort{yolo}v9-e. It is evident that person class is only considered background in some instances, while the helmet is often mistaken with head class. We also show some visual results in the \cref{fig:p_ppe_inf_samples}.

\begin{table}[htbp!]
\centering
\normalsize
\caption{Class-wise accuracy of \acrshort{yolo}v9e model on the Pictor-PPE dataset \cite{nathDeepLearningSite2020}.}
\label{tab:yolo_v9e_inference_on_pictor_ppe}
\normalsize%
\setlength\tabcolsep{3pt}
\begin{tabular}{lrrcccccc}
\toprule
   &  & & & &  \multicolumn{2}{c}{\acrshort{map}} \\
\cline{6-7}
        Class  & Images & Instances &  \acrshort{p} &  \acrshort{r} &   50 &  50-95  \\ \midrule
          all  &      654  &     3477     &  73.5    &  50.3    &  58.9   &   37.6 \\
       person  &      654  &     2080     &  83.6    &  80.7    &  85.5   &   61.6 \\ 
       helmet  &      451  &     1369     &  94.6    &  23.7    &  55.5   &   30.0 \\ 
  safety-vest  &       12  &       28     &  42.4    &  46.4    &  35.5   &   21.2  \\
\bottomrule
\end{tabular}

\end{table}

\begin{figure*}[!htbp]
\input{images/p_ppe_inf/collage_pictor}
\caption{Visual results of inference on Pictor-PPE dataset. Best viewed online. }
\label{fig:p_ppe_inf_samples}
\end{figure*}

\section{Conclusion} \label{sec:conclusion}
Human safety is a critical problem in the manufacturing environment, and in this study, we proposed a human safety dataset for ensuring the \acrshort{ppe} compliance. We have open-sourced 8,099 images annotated in 17 different classes containing different \acrshort{ppe} items and body parts. The dataset is publicly available for research and commercial purposes. To evaluate performance and provide some metrics on the dataset, we trained multiple models for benchmarking, among which \acrshort{yolo}v9-e performed best on the proposed dataset. We evaluated the performance of the trained model with the SH17 dataset on the Pictor-PPE dataset. We obtained satisfactory results that this model can be used in industrial environments to detect and ensure \acrshort{ppe} compliance. For future work, further improvements in the performance of the minority classes with specialized models and custom training methodologies can be considered. 

\section*{Supplementary Materials} Supporting data is available at \href{https://github.com/ahmadmughees/sh17dataset}{SH17 dataset GitHub repository}.

\section*{Authors Contributions}
{\textbf{Hafiz Mughees Ahmad:}
Conceptualization, Methodology, Software, Validation, Data curation, Formal analysis, Visualization, Resources, Writing - Original draft preparation. 
\textbf{Afshin Rahimi:} Funding acquisition, Project administration, Software, Resources, Formal analysis, Supervision, Writing - Review \& Editing. 

\section*{Funding}
This study is supported by IFIVEO CANADA INC., Mitacs through IT16094, \acrfull{nserc} through ALLRP 560406-20, \acrfull{oce} through OCI\# 34166, and the University of Windsor, Canada.

\section*{Institutional Review}
Not applicable.

\section*{Data Availability}
The data presented in this study are available in \href{https://github.com/ahmadmughees/sh17dataset}{SH17 dataset GitHub repository} at \url{https://github.com/ahmadmughees/sh17dataset}.

\section*{Conflict of Interest}{The authors declare that they have no known competing financial interests or personal relationships that could have appeared to influence the work reported in this paper.} 

\section*{Acknowledgement}{A special thanks to Dario Morle, and Syeda Sitara Wishal Fatima from IFIVEO CANADA INC. for their helpful discussions during this study.}

\bibliographystyle{IEEEtran}
\bibliography{IEEEabrv,bib}
\newpage
\appendices
\section{SH17 Dataset}\label{sec:meta-data}
\subsection{Meta Data} \label{subsec:metadata}
We have also provided the collected meta-data of each image that can be used to build the dataset from the source or can be used as additional data during model training. The list of objects and their definitions in the meta data is mentioned in \cref{tab:metadata_format}. We also provide scripts to build the data from the source in the \href{https://github.com/ahmadmughees/sh17dataset}{SH17 dataset Github repository}.
\begin{table}[H]
\centering
\normalsize
\caption{Metadata Format for the Dataset}
\label{tab:metadata_format}
\begin{tabular}{p{4cm}p{9cm}}
\hline
Field & Description \\ \hline
Unique Identifier & A unique code assigned to each image for identification. \\
Width and Height  & The dimensions of the image, specified in pixels. \\ 
URL               & The web address where the image can be accessed. \\ 
Photographer Name & The name of the individual who captured the image. \\ 
Photographer URL  & The web address leading to the photographer's portfolio or profile. \\ 
Photographer ID   & A unique identifier for the photographer, used for reference or database purposes. \\ 
Average Color     & The average color in the hexadecimal code. \\ 
Source            & The origin or platform from which the image was obtained. \\ 
Liked             & Indicates whether the image has been marked as liked or favored on the online platform, typically a boolean value. \\ 
Description       & A summary or narrative about the image's content, context, or theme. \\ \hline
\end{tabular}

\end{table}

\subsection{Person Demographics}
In addition to the bounding boxes representing the Persons class, we have included additional tags based on visible features in the images. While these tags are primarily derived from observable characteristics, they may contain some inaccuracies, as visual features are not always the most reliable indicators. Nonetheless, we include these tags to identify and mitigate potential biases related to gender or ethnic representation. The tags encompass categories such as male, female, children, and various ethnic backgrounds, including Black, Brown, White, and Asian. \Cref{tab:persons_demographics} provides the quantitative analysis of these tags. 
\begin{table}[ht]
\centering
\normalsize
\caption{Quantitative analysis of the additional tags with Persons class.}
\label{tab:persons_demographics}
\begin{tabular}{l|rrr|r}
    \toprule
      \multicolumn{1}{l}{}    & \multicolumn{1}{l}{Male} & \multicolumn{1}{l}{Female} & \multicolumn{1}{l}{Children} & \multicolumn{1}{l}{Total} \\
    \midrule
    White & 2432  & 2032  & 37 & 4501 \\
    Black & 1098  & 776   & 7  & 1881 \\
    Brown & 577   & 218   & 12 & 807 \\
    Asian & 963   & 1272  & 52 & 2287 \\
    \midrule
    Total & 5070  & 4298  & 108& 9476 \\
    \bottomrule
    \end{tabular}

\end{table}

\subsection{Train Test split}
We have used 80\% in train and 20\% in test data. We provide the separate list of training and test files in the \href{https://github.com/ahmadmughees/sh17dataset}{SH17 dataset GitHub repository}. \Cref{tab:sorted_list_of_train_test} provides the complete list of instances in the training and test set. 

\begin{table}[ht]
\centering
\normalsize
\caption{Sorted list of instances in each class in Train and Test Set. }
\label{tab:sorted_list_of_train_test}
\begin{tabular}{rrr}
\toprule
Class             & Train   &Test \\ \midrule
face-guard        &   110   & 24 \\
medical-suit      &   114   & 43 \\
safety-suit       &   195   & 45 \\
ear-mufs          &   269   & 49 \\
safety-vest       &   433   & 97 \\
face-mask-medical &   519   & 151 \\
foot              &   610   & 149 \\
helmet            &   773   & 154 \\
glasses           &   1547   & 398 \\
gloves            &   2261   & 529 \\
shoes             &   3604   & 956 \\
tools             &   3724   & 923 \\
ear               &   6118   & 1612 \\
face              &   7095   & 1855 \\
head              &   9558   & 2427 \\
person            &   11068   & 2734 \\
hands             &   12638   & 3212 \\
\bottomrule
\end{tabular}

\end{table}

\begin{IEEEbiography}[{
\includegraphics[width=1in,height
=1.25in, clip, keepaspectratio]{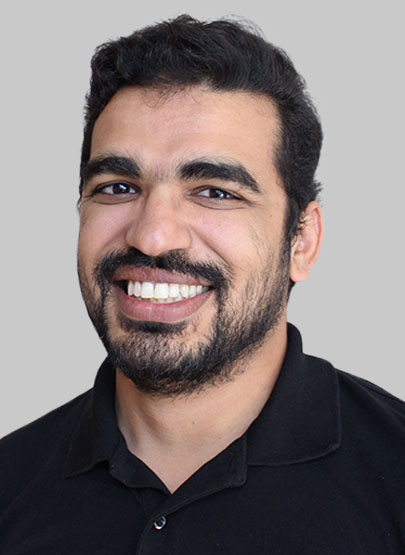}
}]
{Hafiz Mughees Ahmad} completed his Bachelor's and Master's in Electrical Engineering from the Institute of Space Technology, Pakistan, in 2015 and 2018, respectively. He is currently pursuing a Ph.D. at the University of Windsor, Canada. Alongside his studies, he serves as a Deep Learning Engineer at IFIVEO CANADA INC. His previous roles include Research Associate at Istanbul Medipol University, Turkey, and Lecturer at the Institute of Space Technology, Pakistan. His research focuses on Computer Vision and Deep Learning, with applications in OD and real-time surveillance and monitoring in industrial manufacturing and production environments. He is a Graduate Student Member of IEEE.
\end{IEEEbiography}
\vskip -2\baselineskip plus -1fil
\begin{IEEEbiography}
[{
\includegraphics[width=1in,height=1.25in,clip,keepaspectratio]{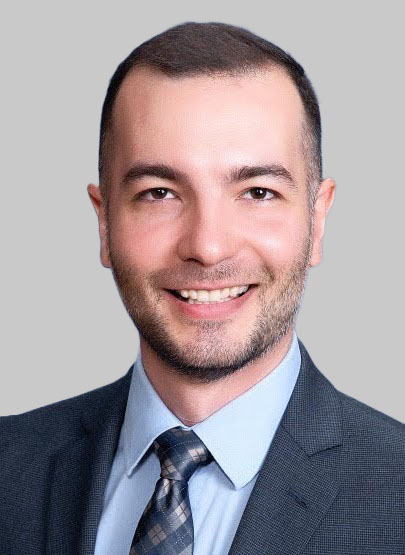}
}]
{Afshin Rahimi} received his B.Sc. degree from the K. N. Toosi University of Technology, Tehran, Iran, in 2010, and the M.Sc. and Ph.D. degrees from Toronto Metropolitan University, Toronto, ON, Canada, in 2012, and 2017, respectively, in Aerospace Engineering. He was with Pratt \& Whitney Canada from 2017 to 2018. He is currently an Associate Professor in the Department of Mechanical, Automotive, and Materials Engineering at the University of Windsor, Windsor, ON, Canada. Since 2010, he has been involved in various industrial research, technology development, and systems engineering projects/contracts related to the control and diagnostics of satellites, UAVs, and commercial aircraft subsystems. In recent years, he has also been involved with industrial automation and using technologies to boost manual labor work in industrial settings. He is a senior member of IEEE, a lifetime member of AIAA, and a technical member of the PHM Society.
\end{IEEEbiography}
\vskip -2\baselineskip plus -1fil

\end{document}